\documentclass[UTF8]{article} 
\usepackage{iclr2020_conference_modify}
\usepackage{times}

\usepackage{amsmath,amsfonts,bm}

\def\eqref#1{equation~\ref{#1}}

\def\1{\bm{1}}

\DeclareMathAlphabet{\mathsfit}{\encodingdefault}{\sfdefault}{m}{sl}
\SetMathAlphabet{\mathsfit}{bold}{\encodingdefault}{\sfdefault}{bx}{n}

\usepackage{hyperref}
\usepackage{url}
\usepackage{natbib}
\usepackage{graphicx}

\title{Towards a Unified Evaluation of Explanation Methods without Ground Truth}

\author{Hao Zhang, Jiayi Chen, Haotian Xue, Quanshi Zhang\thanks{Quanshi Zhang is the corresponding author, \texttt{zqs1022@sjtu.edu.cn}. This work was done when Hao Zhang, Jiayi Chen and Haotian Xue were research interns at Quanshi Zhang's lab.}\\
Shanghai Jiao Tong University\\
}

\newcommand{\tabincell}[2]{\begin{tabular}{@{}#1@{}}#2\end{tabular}}

\iclrfinalcopy 
\begin{document}

\maketitle

\begin{abstract}
This paper proposes a set of criteria to evaluate the objectiveness
of explanation methods of neural networks,
which is crucial for the development of explainable AI,
but it also presents significant challenges.
The core challenge is that people usually cannot obtain ground-truth explanations of the neural network.
To this end, we design four metrics to evaluate explanation results without ground-truth explanations.
Our metrics can be broadly applied to nine benchmark
methods of interpreting neural networks,
which provides new insights of explanation methods.
\end{abstract}

\section{Introduction}

\label{sec-intro}
Nowadays, many methods are proposed to explain the feature representations of a deep neural network (DNN)
in a post-hoc manner.
In this research, we limit our attention to existing methods of estimating the
\emph{importance/attribution/saliency} of input pixels or
intermediate-layer neural units
\emph{w.r.t.} the network output~\citep{shrikumar2016not, lundberg2017unified,
ribeiro2016should, binder2016layer},
which present the mainstream of explaining neural networks.
To avoid ambiguity, the estimated importance/saliency/attribution maps are all termed ``attribution maps" in this paper.

However, some methods usually pursue attribution maps which look reasonable
from the perspective of human users,
instead of objectively reflecting the information processing in the DNN.
A trustworthy evaluation of the objectiveness of attribution maps is crucial for the
development of deep learning and proposes significant challenges to state-of-the-art algorithms.

Existing metrics~\citep{yang2019bim, arras2019evaluating}
of evaluating explanation methods have certain shortcomings.

\textbf{Issue 1, evaluation of the accuracy of a DNN $\not =$
evaluation of the objectiveness of attribution maps:}
\emph{
Some methods only evaluate whether the visualized attribution map looks reasonable to human users,
instead of examining whether an attribution map objectively reflects the truth of a DNN.}
~\citep{cui2019integrative, yang2019evaluating}
used human cognition to evaluate the explanation result.
~\citep{yang2019bim, kim2017interpretability, camburu2019i} aimed to construct a specific dataset
with ground-truth explanations for evaluation.
For example, they added an irrelevant object into the image.
Pixels from the irrelevant object are expected to be assigned with zero attributions.

However, strictly speaking, it is impossible to religiously annotate ground-truth explanations for a DNN.
Currently, the ground-truth explanation is constructed under the assumption that a DNN cannot learn irrelevant objects
for classification.
Its purpose was to evaluate attribution maps of the DNN,
instead of examining whether an explanation method mistakenly generates
seemingly correct attribution maps which do not reflect the truth of an incorrectly learned DNN.

\textbf{Issue 2, broad applicability:}
We aim to design an evaluation metric that can be broadly applied to various tasks.
In aforementioned methods~\citep{yang2019bim, kim2017interpretability, camburu2019i},
the requirement for constructing a new testing dataset limits the applicability of the evaluation.

\textbf{Issue 3, quantification of the objectiveness:}
Some methods quantitatively evaluate the accuracy and robustness of attribution maps.
However, there is no strict mechanism to ensure the objectiveness
of each numerical value in the attribution map.
\emph{I.e.}, if the attribution value of a pixel is twice of that of another pixel,
then the first pixel is supposed to contribute twice numerical values to the prediction
\emph{w.r.t.} the second pixel.

Except for the objectiveness, previous studies also conducted the evaluation from other perspectives.
~\citep{arras2019evaluating, vu2019evaluating} evaluated attribution maps from the perspective of adversarial
attacks by adding random noise.
~\citep{adebayo2018sanity, ghorbani2019interpretation, David2018Towards}
proposed methods to evaluate the robustness of explanation methods \emph{w.r.t.} the perturbation.
~\citep{adebayo2018sanity} randomized the layer of DNN from the top to the bottom
and visualized the change of attribution maps.

\begin{table*}[!t]
\caption{Review of explanation methods}
\label{method-conclusion-table}
\begin{center}
\resizebox{0.9\linewidth}{!}{\
\begin{small}
\begin{tabular}{cp{0.2\linewidth}p{0.4\linewidth}}
\hline
Method & What to explain & Qualitative evaluation of limitations in application\\
\hline
\tabincell{c}{CAM~\citep{zhou2016learning} }   & Attribution distribution at intermediate layer
&1. Requirement for global average pooling. 2. Usually explain features at high layers\\
\hline
\tabincell{c}{Grad\_CAM~\citep{selvaraju2017grad}} & Attribution distribution at intermediate layer &
Usually explain features at high layers\\
\hline
\tabincell{c}{Grad~\citep{simonyan2013deep}} & Pixel-wise attribution & --\\
\hline
\tabincell{c}{GI~\citep{shrikumar2016not}}  & Pixel-wise attribution & --\\
\hline
\tabincell{c}{GB~\citep{springenberg2014striving}}  & Pixel-wise attribution &
Requirement for using ReLU as non-linear layers\\
\hline
\tabincell{c}{Shapley Value~\citep{shapley1953value}} & Pixel-wise attribution & NP-complete problem\\
\hline
\tabincell{c}{DeepSHAP~\citep{lundberg2017unified}} &
Pixel-wise attribution & Similar to LRP, DeepLIFT~\citep{shrikumar2016not} with a designed backward rule\\
\hline
\tabincell{c}{LIME~\citep{ribeiro2016should}} & Pixel-wise attribution &  Attribution maps at the super-pixel level,
rather than at the pixel level\\
\hline
\tabincell{c}{LRP~\citep{binder2016layer}} & Pixel-wise attribution & Relevance propagation rules of every layer should be defined\\
\hline
\tabincell{c}{Pert~\citep{fong2017interpretable}}& Pixel-wise attribution & -\\
\hline
\end{tabular}
\end{small}
}
\end{center}
\vspace{-15pt}
\end{table*}

Considering the above three issues, in this study,
we aim to fairly evaluate attribution maps generated by explanation methods without ground truth.
The evaluation of attribution maps needs to be conducted from the following four perspectives,
which are supplementary to each other, \emph{i.e.}, objectiveness, completeness, robustness, and commonness.
First, many explanation methods usually trade off between objectiveness and completeness, \emph{i.e.},
choosing either to objectively report accurate attributions of salient (easy) regions,
or to pay more attention to non-salient (difficult) regions for more complete explanation.
On the other hand, we also analyze the robustness and commonness of explanations.
Common explanations shared by different explanation methods are usually robust in real applications,
although common explanations do not always objectively reflect the truth of a DNN.
Note that in most applications, people cannot faithfully obtain ground-truth attribution maps.
Therefore,the objectiveness, completeness, robustness, and commonness of explanation methods need to be evaluated without ground-truth,
which is the distinct contribution compared to previous studies.

\textbf{1. Objectiveness, bias of the attribution map at the pixel level:}
In order to evaluate the bias of the attribution map,
we first need to propose a standard metric to evaluate the accuracy of explanation methods.
The Shapley value is the unique solution to model the
attribution value of each pixel that satisfies desirable properties including
efficiency, symmetry and monotonicity~\citep{lundberg2017unified}.
However, the computation of the Shapley value is an NP-complete problem,
and previous studies~\citep{lundberg2017unified} showed that the accurate estimation
of the Shapley value is still a significant challenge.
To this end, we extend the theory of Shapley sampling~\citep{castro2009polynomial} and design a
new evaluation metric, which achieves high accuracy without significantly boosting the computational cost.

We use the new evaluation metric to quantify the bias of the attribution map.
Note that this evaluation has no partiality to the Shapley-value-based
explanation methods.
For example, experimental results showed that LRP~\citep{binder2016layer}
exhibited significantly lower bias than DeepSHAP~\citep{lundberg2017unified}.

\textbf{2. Completeness, quantification of unexplainable feature components:}
Given an input image and its attribution map,
we revise the input image to generate a new image in which we mask unimportant regions.
We then compare the intermediate-layer feature of the original image with that of the generated image,
so as to disentangle feature components that can and cannot be explained by the attribution map.

\textbf{3. Robustness, robustness of the explanation:} Robustness of the explanation means whether the
attribution map is robust to spatial masking of the input image.
When we randomly mask a certain region of the input image,
we admit that spatial masking destroys global contexts and affects pixel-wise attribution value
to some extent.
The quantification of the robustness of the explanation is an important perspective of evaluating an explanation method.

\textbf{4. Commonness, mutual verification:} The mutual verification means whether different explanation
methods can verify each other.
Methods generating similar attribution maps are usually believed more reliable.

In this paper, we used our metrics to evaluate
nine widely used explanation methods listed in Table~\ref{method-conclusion-table}.
We conducted experiments using the LeNet, VGG and ResNet on different benchmark datasets
including the CIFAR-10~\citep{krizhevsky2009learning} dataset and the Pascal VOC 2012~\citep{Everingham10} dataset.
Our experimental results proved the effectiveness of the proposed evaluation methods and provided an insightful
understanding of various explanation methods.

The contribution of our work can be concluded as follows.\\
\noindent$\bullet\quad$In this study, we invent a set of standard metrics to evaluate the objectiveness and the robustness of
the attribution map without knowing ground-truth explanations.\\
\noindent$\bullet\quad$The metric of evaluating the pixel-wise bias of the attribution map can be estimated
with a relatively low computational cost,
which avoids falling into the computational bottleneck of estimating accurate pixel-wise attributions.\\
\noindent$\bullet\quad$Since our metrics do not need any annotations of ground-truth explanations,
our metrics can be applied to different neural networks trained on different datasets.

\section{Related Work}
\label{sec-rel}
Explainable AI is an emerging direction in artificial intelligence,
and different explanation methods have been proposed.

Firstly, the visualization of feature representations inside a DNN is the most direct way of opening the
black-box of the DNN.
Related techniques include gradient-based visualization
~\citep{CNNVisualization_1,CNNVisualization_2,CNNVisualization_6} and up-convolutional nets~\citep{FeaVisual}.
Secondly, other studies diagnose feature representations inside a DNN.
~\citep{patternNet} extracted rough pixel-wise correlations between network inputs and outputs,
\emph{i.e.}, estimating image regions with large influence on the network output.
Network-attack methods~\citep{CNNInfluence,CNNAnalysis_1} computed adversarial samples to diagnose a CNN.
Bau \emph{et al.}~\citep{Interpretability} defined six types of semantics for CNN filters, \emph{i.e.} objects, parts, scenes,
textures, materials, and colors.
Fong and Vedaldi~\citep{net2vector} analyzed how multiple filters jointly represented a certain visual concept.
Thirdly, a recent new trend is to learn interpretable features
in DNNs~\citep{LogicRuleNetwork,CNNCompositionality,Parsimonious}.
Capsule nets~\citep{capsule} and interpretable RCNN~\citep{InterRCNN} learned interpretable features in
intermediate layers.
InfoGAN~\citep{infoGAN} and $\beta$-VAE~\citep{betaVAE} learned well-disentangled codes for generative neural networks.

However, to simplify the story, in this paper, we briefly review the Shapley value and limit our discussions to existing methods of
evaluating methods of extracting attribution/importance/saliency maps.

\textbf{The Shapley value:}
The Shapley value~\citep{shapley1953value} was proposed
to compute the attribution distribution over all players in a
particular cooperative game.
However, it is an NP-complete problem to compute the accurate Shapley value.
The Shapley value approximated by sampling strategy could be very inaccurate due to the high variance.
We extend the theory of the Shapley value to obtain an evaluation metric with a high accuracy
but a low computational cost.

\textbf{Qualitative evaluation:}
Some studies used a qualitative criterion for evaluation.
~\citep{cui2019integrative} qualitatively defined basic concepts in the evaluation of explanation results,
including the complexity of the explanation, the correlation, and the completeness.
~\citep{yang2019evaluating} qualitatively evaluated explanation methods according to their generalizability,
fidelity and persuasibility.
In contrast, this paper aims to evaluate the methods
quantitatively, which makes our metrics more objective and reliable.

\textbf{Accuracy evaluation:}
To evaluate the accuracy of attribution maps, ~\citep{arras2019evaluating, vu2019evaluating}
used the noise/occlusion to perturb the original image according to the attribution value.
However, there was no mechanism to ensure the prediction result objectively reflected the truth
of a DNN.
~\citep{yang2019bim, kim2017interpretability} built a dataset to help them generate ground-truth explanations.
Essentially, these methods tried to evaluate the correctness of attributions.
However, a rigorous study should not assume that the DNN can make inference in the same way as people.
~\citep{OramasVisual} proposed four metrics to evaluate the explanation from four different perspectives,
which can provide a comprehensive understanding towards explanation methods.
However,~\citep{OramasVisual} also used synthetic datasets and prediction results for evaluation.
As a result, this paper proposes to evaluate the objectiveness of explanation results
without annotations of ground-truth explanations.

\textbf{Stability evaluation:}
~\citep{adebayo2018sanity, ghorbani2019interpretation, David2018Towards} mainly paid attention to the
attribution map change when the model input was perturbed.
~\citep{adebayo2018sanity} visualized the change in the attribution map when the weights of the model were
destroyed from the top to the bottom.
~\citep{ghorbani2019interpretation, David2018Towards} used the adversarial image to alter the attribution map.
In comparison, we propose a metric to evaluate the robustness to spatial masking.

\section{Algorithm}
\label{sec-alg}

\subsection{Preliminaries: The Shapley value}
\label{subsec-shapley}
The Shapley value measures the instancewise feature importance ranking problem.
Let $\Omega$ be the set of all pixels of an image $I$.
$I_P$ denotes an image that replaces all pixels in set $\Omega\setminus P$ with average pixel value over images.
$F(I_P)$ denotes the scalar output of a DNN based on a
subset of pixels $P \subset \Omega$.
To compute the Shapley value of the $i$-th feature,~\citep{shapley1953value} considered all subsets of
$\Omega$ not containing the $i$-th feature and defined the Shapley value $A^{*}_i$ as follows:
\begin{small}
\begin{equation}
\label{equ:cls-SHAP}
A^{*}_i = \sum_{P\subset \Omega\setminus \{i\}}\frac{|P|!(|\Omega|-|P|-1)!}{|P|!}
\left[F(I_{P\cup \{i\}})- F(I_P)\right]
\end{equation}
\end{small}
It is the unique solution that satisfies several desirable properties to assign attribution value to each
feature dimension in the input,
which have been well introduced in~\citep{Chen2018L}.
For convenience of the reader, we also summarize such properties in~\citep{Chen2018L} in the supplementary material.

\subsection{Evaluating the bias of the attribution map at the pixel level}
\label{subsubsec-shapeval}
In this section, we design a metric to accurately evaluate the objectiveness of the attribution map.
Given an image $I \in \bf I$, let us consider the DNN $F$ with a single scalar output $y=F(I)$.
For DNNs with multiple outputs, existing methods usually explain each individual
output dimension independently.
Let $\{a_i\}$ denote the pixel-wise attribution map estimated
by a specific explanation method.
We aim to evaluate the bias of $\{a_i\}$.
People usually formulate the network output as the sum of pixel-wise attribution value,
\emph{i.e.} the output $y$ can be decomposed as follows.
\begin{small}
\begin{equation}
\label{equ:additive}
y = b + \sum_{i\in \Omega}A_i,\qquad \textrm{s.t.}\quad A_i=\lambda a_i
\end{equation}
\end{small}
$b$ denotes the bias; $i$ denotes the index of each pixel in the input image;
$\Omega$ denotes the set of all pixels in the image.
Aforementioned $\{A^*_i\}$ can be considered as the ground-truth of $\{A_i\}$~\citep{lundberg2017unified}.
Since many explanation methods~\citep{selvaraju2017grad, simonyan2013deep} mainly compute
relative values of attributions $\{a_i\}$,
instead of a strict attribution map $\{A_i\}$.
We use $\lambda$ to bridge $\{A_i\}$ and $\{a_i\}$.
$\lambda$ is a constant for normalization,
which can be eliminated during the implementation of the evaluation.

The estimated attribution of each pixel can be assumed to follow a Gaussian distribution
{\small $A_i\sim \mathcal{N}(\mu_i,\sigma^2_i)$}~\citep{castro2009polynomial}.
Attribution distributions of different pixels can be further assumed to share a unified variance, \emph{i.e.}
$\sigma^2_1\approx\sigma^2_2\approx...\approx\sigma^2_n$.
The evaluation of the attribution distribution $\{a_i\}$ has two aspects, \emph{i.e.}\\
1. the sampling of pixels whose attributions are more likely to have large deviations;\\
2. the evaluation of the bias of the sampled attributions.

\begin{figure}[!t]
	\begin{center}
	\includegraphics[width=0.6\linewidth]{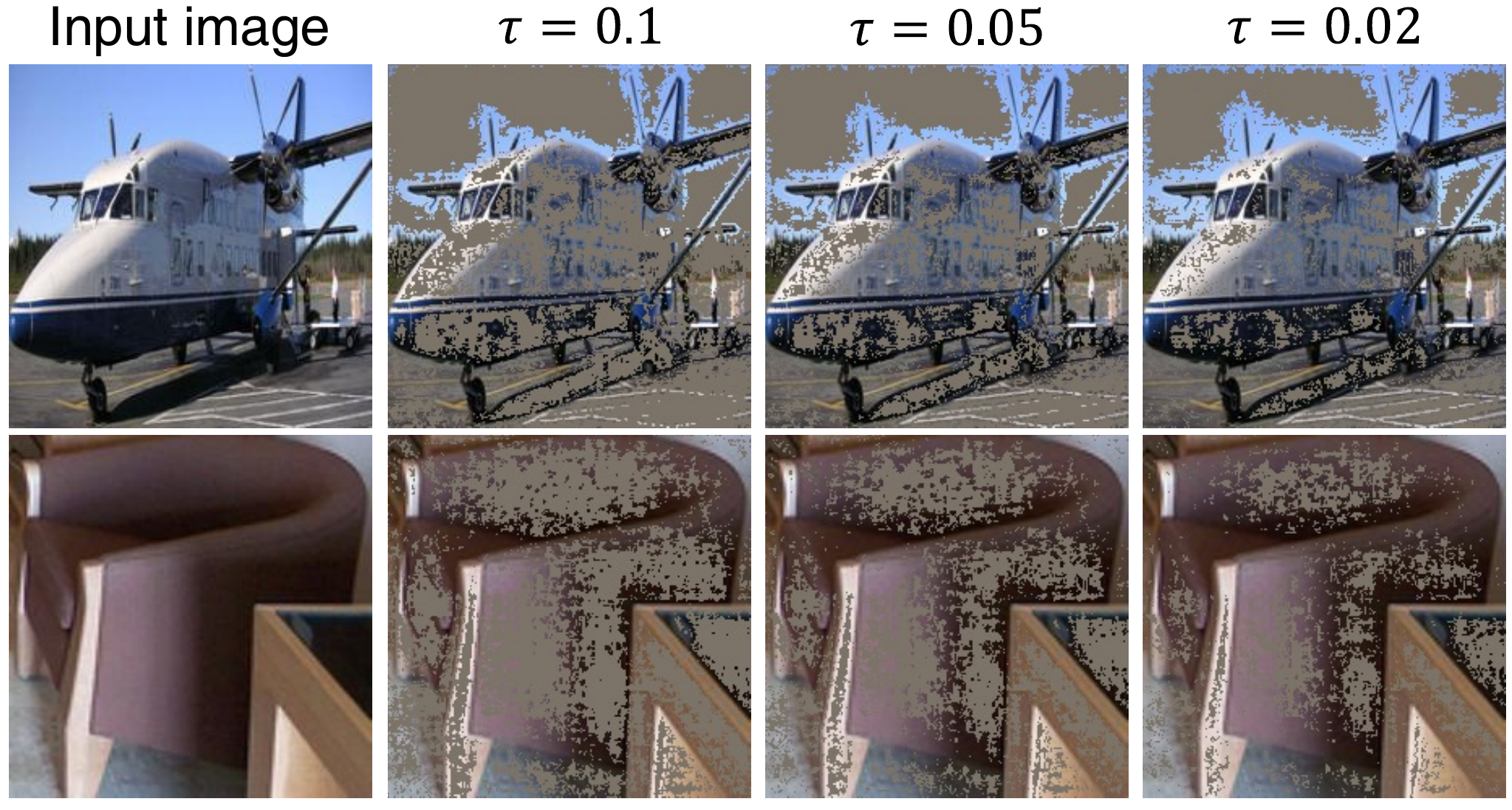}
	\end{center}
	\caption{Masked image regions with different value of $\tau \,(0<\tau<1)$.
    We need to set a small value of $\tau$ to make sure that unmasked image regions contain most attributions.
    When we set $\tau$ to 0.02, 0.05 and 0.1, the small difference of masked images doesn't significantly affect conclusion of the evaluation.
    Thus, we set $\tau=0.05$ in this paper.}
	\label{fig:dif-tau}
    \vspace{-5pt}
\end{figure}

First, for the sampling of attributions of interest,
we sample the set of pixels $S$ with top-ranked high (or low) attributions.
Attribution values of pixels in $S$ are sampled as those with the highest (or the lowest) values,
and these pixels are supposed to be more likely to be significantly biased
towards high (or low) attribution values.
Meanwhile, from another perspective,
the distribution of the sampled attribution values is close to the Gumbel distribution.

Second, although the Shapley value can be considered as a standard formulation of the pixel-wise attribution,
it usually cannot be accurately computed because of its high computational cost.
In order to accurately evaluate the sampled attribution values without significantly increasing the computational cost,
we applied the Shapley value approximated by the sampling method.
Just like the target attribution distribution $A_i$,
the approximated Shapley value $A_i^{shap}$ is assumed to follow a normal distribution
{\small $\mathcal{N}(A_i^{*}, (\sigma^{shap})^2)$}.
$A_i^{shap}$ is an unbiased approximation of the true Shapley value $A_i^{*}$,
Thus, the average value over different pixels in $S$ satisfies
{\small $\frac{\sum_{i\in S} A_i^{shap}}{|S|}\sim \mathcal{N}(\frac{\sum_{i\in S}A_i^{*}}{|S|},
\frac{(\sigma^{shap})^2}{|S|})$}.

We can prove that the measurement of the average attribution among all sampled pixels
{\small $\frac{\sum_{i\in S}A_i^{shap}}{|S|}$}
is of much higher accuracy
than the raw Shapley value with the same computational cost.
The difference between the highest (or lowest) values and its true values
{\small $\Big|\frac{\sum_{i\in S}A^{shap}_i}{|S|\|A^{shap}\|}-\frac{\sum_{i\in S}A_i}{|S|\|A\|}\Big|$}
can reflect the system bias, as follows.
\begin{small}
\begin{equation}
\begin{aligned}
\label{equ:shap-ratio}
M_{\textrm{pixel}}&=\mathbb{E}_{I}\left[\Big|\frac{\sum_{i\in S}A^{shap}_i}{|S|\|A^{shap}\|}-
    \frac{\sum_{i\in S}A_i}{|S|\|A\|}\Big|\right]\\
    &=\mathbb{E}_{I}\left[\frac{1}{|S|}\Big|\frac{\sum_{i\in S}A^{shap}_i}{\|A^{shap}\|}-
    \frac{\sum_{i\in S}a_i}{\|a\|}\Big|\right] \\
    &\textrm{s.t.}\qquad \forall i\in S, j\in\Omega\setminus S, a_i\ge a_j \quad \\
    &\quad \quad \textrm{or} \quad
\forall i\in S, j\in\Omega\setminus S, a_i\le a_j
\end{aligned}
\end{equation}
\end{small}

where $\Omega$ is the set of all pixels in an image.
$\|A^{shap}\|$ and $\|a\|$ are used for normalization.
A small value of $M_{\textrm{pixel}}$ indicates the low bias of the attribution map.

\textbf{Analysis of the high computational efficiency:}
Suppose that the computational complexity of processing one sample is {\small $\mathcal{O}(N)$},
then the computational complexity of sampling $m$ times is {\small $\mathcal{O}(mN)$}.
If the raw Shapley value needs to obtain the same accuracy,
it needs significantly more samples, and the computational complexity is {\small $\mathcal{O}(|S|mN)$}.
Please see the supplementary material for further discussions about the computational cost.

In addition, the proposed metric can also be used to evaluate the attribution of neural
activations in the intermediate layer,
such as those generated by Grad-CAM.
In this case, we can regard the target intermediate-layer feature as the input image to
compute attributions,
so as to implement the evaluation.
For each image, we need to sample multiple times to increase the accuracy.
We compute the average performance over different images for evaluation.
We need to sample multiple times with different images to increase the accuracy of the evaluation.
Note that although the metric is designed based on the Shapley value,
experimental results showed that LRP outperforms DeepSHAP.

\begin{figure*}[!t]
    \centering
    \includegraphics[width=0.9\linewidth]{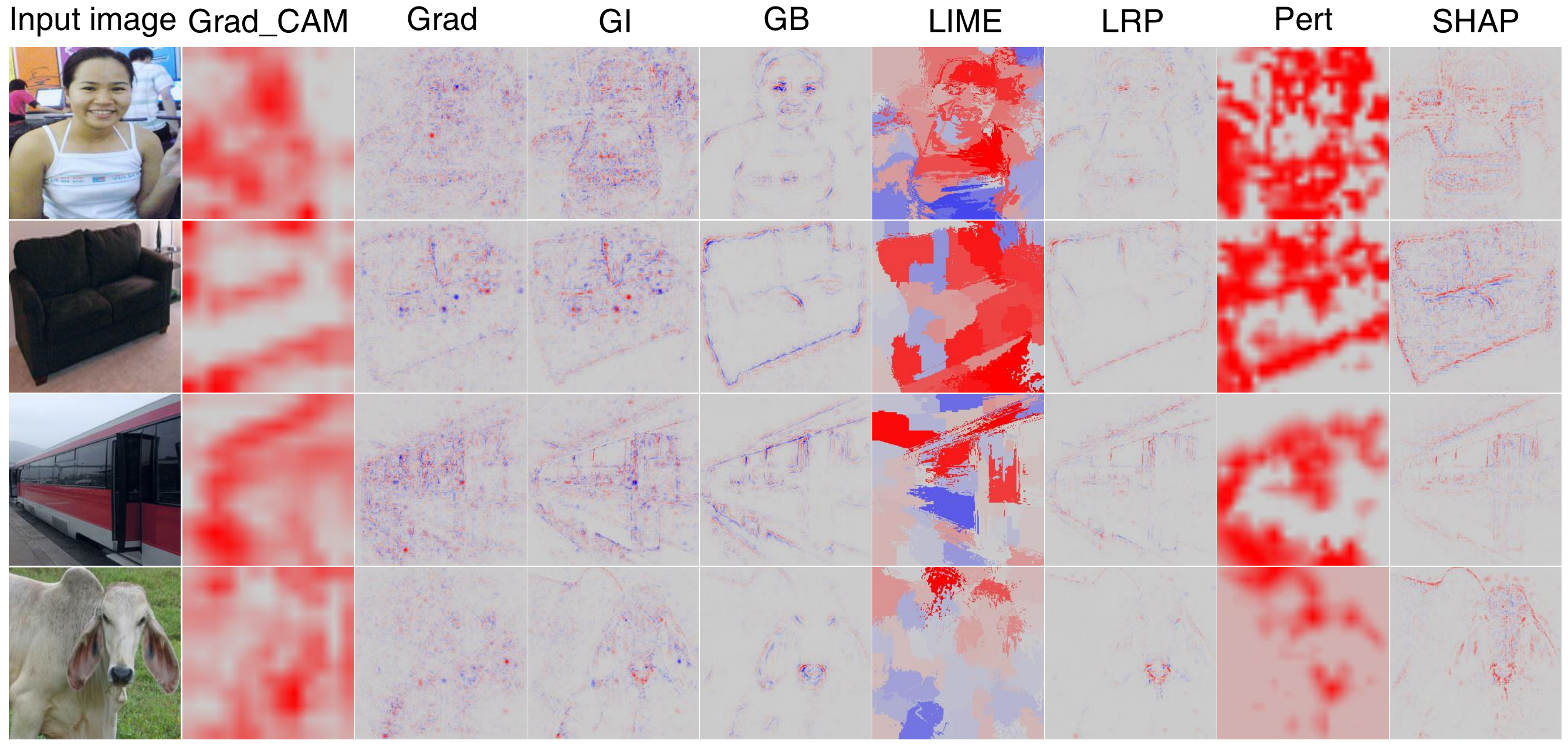}
    \caption{Examples of attribution maps of different methods. The supplementary material provides more attribution maps.}
    \vspace{-5pt}
    \label{fig:attribution}
\end{figure*}

\subsection{Quantification of unexplainable feature components}
\label{subsubsec-decoder}
We propose another metric to quantify unexplainable feature components.
Given an image $I$ and its attribution map $\{a_i\}$, we generate a new image,
which reflects the reasoning of the attribution map.
In this way, we can consider the feature of the newly generated image $\widetilde{f}_{I}$
as feature components that can be explained.
Let $f_{I}$ denote the feature of the original image $I$.
Then, $f_{I}-\widetilde{f}_{I}$ corresponds to the unexplainable feature components.

To generate the new image, we mask specific pixels in the original image $I$,
which have the lowest attributions.
We select and mask a set of pixels $S$ with the lowest absolute attributions,
and the number of the selected points is determined subjects to
$\sum_{i\in S}|a_i|=\tau \sum_{i\in \Omega}|a_i|$ to generate the new image $\widetilde{I}$.

Note that $\tau$ is a small positive scalar to control the size of mask.
In this study, we choose the value of $\tau$ as $0.05$.
As shown in the Figure~\ref{fig:dif-tau},
we set $\tau=0.05$, \emph{i.e.}, we mask pixels with lowest attributions.
The masked pixels only make 5\% attributions in total.
In real application, the conclusion is not sensitive to the value of $\tau$.
Promising evaluation results obtained with $\tau=0.05$ is similar to those obtained with $\tau=0.1$.
(Please see the supplementary material for results with $\tau=0.1$.)
The metric is formulated as
\begin{small}
\begin{equation}
    \label{equ:decoder}
    M_{\textrm{feature}}=\alpha \mathbb{E}_{I}\left[\|\widetilde{f}_{I} - f_{I} \|\right]
\end{equation}
\end{small}
where $\alpha = \frac{1}{\mathbb{E}_{I'}[\|f_{I'}-\mathbb{E}_{I''}[f_{I''}]\|]}$ is used for normalization.
A small value of $M_{\textrm{feature}}$ indicates most feature components in $f$ are explainable.

\begin{figure*}[!t]
    \centering
    \includegraphics[width=0.99\linewidth]{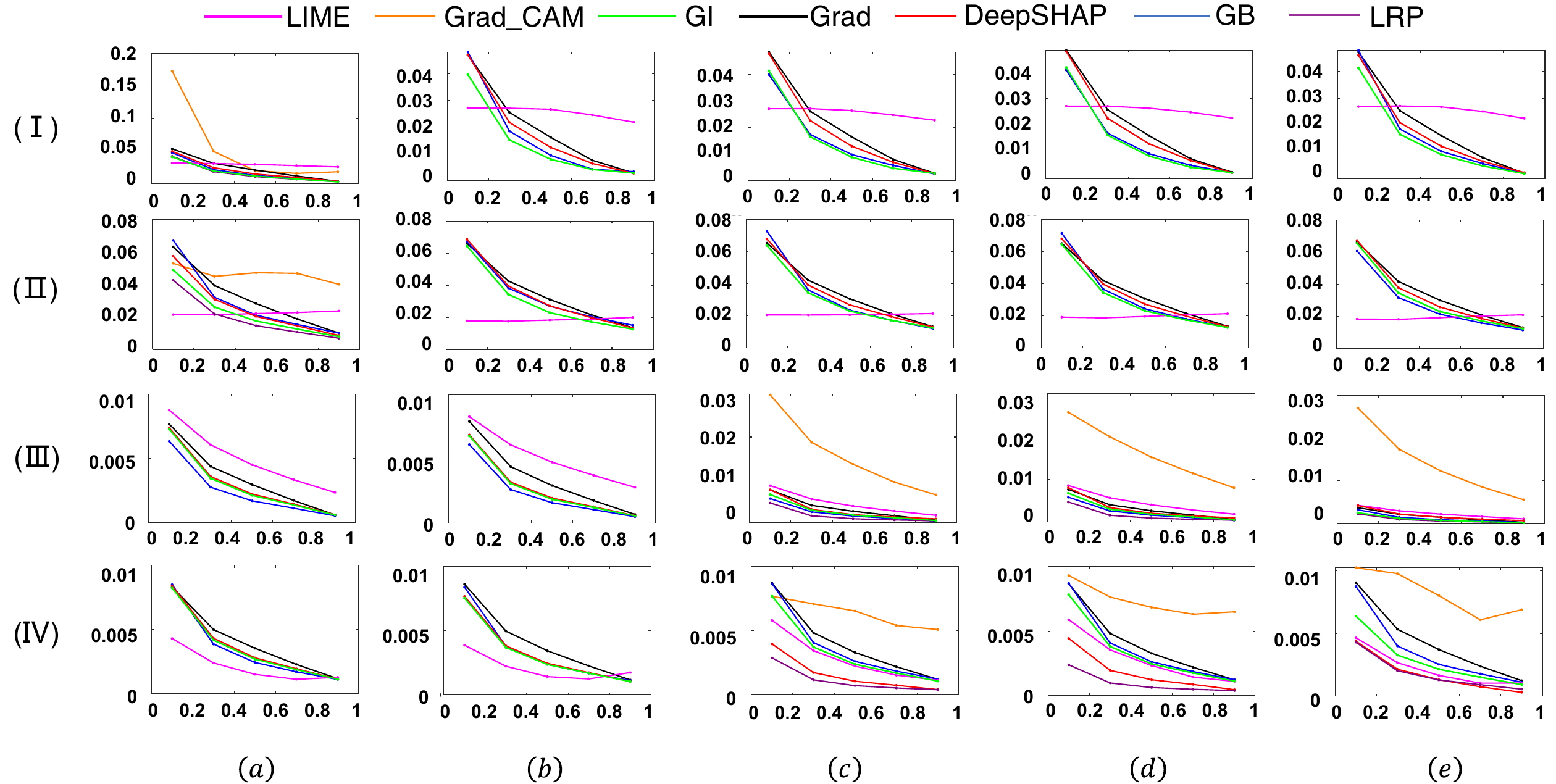}
    \vspace{5pt}
    \caption{Results of the bias of the attribution map at the pixel level.
    Row (\uppercase\expandafter{\romannumeral1}) and row (\uppercase\expandafter{\romannumeral2})
    used trained LeNet, ResNet20, ResNet32, ResNet44, ResNet56 on the CIFAR-10 dataset from left to right;
    row (\uppercase\expandafter{\romannumeral3}) and row (\uppercase\expandafter{\romannumeral4})
    used trained ResNet50, ResNet101, VGG16, VGG19, AlexNet on the Pascal VOC 2012 dataset from left to right.
    Row (\uppercase\expandafter{\romannumeral1}) and row (\uppercase\expandafter{\romannumeral3})
    sampled pixels with the highest attribution values;
    row (\uppercase\expandafter{\romannumeral2}) and row (\uppercase\expandafter{\romannumeral4})
    sampled the pixels with the lowest attribution values.
    Explicit numbers in these curves are listed in the supplementary material.}
    \vspace{-5pt}
    \label{fig:pixel-level}
\end{figure*}

\subsection{Evaluating the robustness of explanation}
\label{subsec-stable}
This metric is used to measure the robustness of explanation methods to the spatial masking.
We believe that the method, which is robust to spatial masking, can be considered more convincing.
The robustness is an important perspective of evaluating explanation methods.

Given an input image $I \in \bf I$ and its attribution map $\{a_i\}$ \emph{w.r.t} a DNN,
we use a mask $M$ to cover specific parts of the image to get a masked image $\hat{I}$.
For each input image $I$, we can generate four masked images by masking
the right, left, top, and bottom half of the image, respectively.
For each masked image $\hat{I}$, the explanation method estimates the attribution $\hat{a}_i$ for each pixel.
Note that for the masked image, we still compute attributions
with respect to target classification result for fair comparisons.
We compare pixel-level attributions of the unmask pixels between original images and masked images, as follows.
\begin{small}
\begin{equation}
    \label{equ:stability}
    M_{\textrm{non-robust}}=\mathbb{E}_{I}\left[\frac{1}{\|a\|}\sqrt{\sum_{i\in I\setminus I_{mask}}
    (a_i-\hat{a}_i)^2}\right]
\end{equation}
\end{small}
We used $\|a\|$ for normalization,
and a large value of $M_{\textrm{non-robust}}$ indicates a high non-robustness.

\subsection{Evaluating the mutual verification}
\label{subsec-auth}
This metric aims to quantitatively measure the mutual verification between different explanation methods.
Adebayo \emph{et al.}~\citep{adebayo2018sanity} showed that some methods generated similar explanations,
but they were all unreliable and biased.
However, such a metric still provided some new insights on the relationship between different explanation methods.
Given a DNN $F$ and an image $I\in \bf I$,
two different explanation methods $\alpha$ and $\beta$
produce attribution maps $a_\alpha$ and $a_\beta$, respectively.
We measure their difference as follows.

\begin{small}
\begin{equation}
    \label{equ:auth}
    M_{\textrm{mutual}}=\mathbb{E}_{I}\left[\|\frac{a_\alpha}{\Vert a_\alpha\Vert}-\frac{a_\beta}{\Vert a_\beta\Vert}\|\right]
\end{equation}
\end{small}

Attribution maps from different explanation methods are normalized by their L2-norm.
A lower value of $M_{\textrm{mutual}}$ indicates a more convincing mutual verification
between explanation methods $\alpha$ and $\beta$.

\subsection{Discussion about Limitations}
\label{subsec-limit}
Table~\ref{table:limitation} shows the limitation and applicability of these metrics.
Note that different applications may have their own evaluation metrics.
Nevertheless, in this paper, we focus on some common desirable properties that need to be shared by evaluation metrics,
\emph{i.e.} the objectiveness, completeness, robustness, and commonness.
They provide new insights on explanation methods.

\begin{table}[!t]
\caption{Limitations of different evaluation metrics.}
\begin{center}
\resizebox{0.99\linewidth}{!}{\
\begin{small}
\begin{tabular}{p{0.33\linewidth}|p{0.66\linewidth}}
\hline
Metrics & Limitation \& Applications \\
\hline
Bias of the attribution map & The method must calculate an attribution map with both positive and negative values.  \\
\hline
Quantification of unexplainable feature components & The attribution map must be computed for pixels in the input image
instead for units in the intermediate-layer feature. \\
\hline
Robustness of explanation method & We only evaluate the robustness towards spatial masking. There can be other kinds of robustness. \\
\hline
Mutual verification & The compared attribution maps must be calculated for the same item (\emph{e.g.} input image or intermediate feature). \\
\hline
\end{tabular}
\end{small}
}
\end{center}
\vspace{-5pt}
\label{table:limitation}
\end{table}

\begin{table*}[!t]
\caption{Quantification of unexplainable feature components with $\tau=0.05$.
LIME usually outperformed other explanation methods.}
\begin{center}
\resizebox{0.9\linewidth}{!}{\
\centering
\begin{tabular}{l|ccccccc}
Method & Grad& GI & GB & DeepSHAP & LIME & LRP & Pert \\
\hline
CIFAR10-LeNet &	0.82971	&	0.98173	& 0.81278	& 0.98706 & \bf0.54371	&	0.97856	&   0.58968	\\
CIFAR10-ResNet20 &	1.11988	&	1.21033	&	1.09124	&	1.20117	&	\bf0.93836	&	-	&	1.05067\\
CIFAR10-ResNet32 &	1.16659	&	1.22853	&	1.12193	&	1.20667	&	\bf0.91767	&	-	&	1.07155\\
CIFAR10-ResNet44&	1.13872	&	1.19983	&	1.03684	&	1.1786	&	\bf0.89585	&	-	&	1.01908\\
CIFAR10-ResNet56 &	1.12963	&	1.18625	&	1.05485	&	1.17067	&	\bf0.89013	&	-	&	1.02354\\
VOC2012-AlexNet &	0.91919	&	0.99364	&	0.97473	&	0.98271	&	\bf0.89007	&	1.02604	&	0.9409	\\
VOC2012-VGG16 & 1.11413	&	1.13402	&	1.16247	&	1.23222	&	\bf0.98048	&	1.25679	&	1.01887 \\
VOC2012-VGG19 &	1.00716	&	1.06308	&	1.04998	&	1.11102	&	\bf0.95401	&	1.17729	&	1.02434	\\
VOC2012-ResNet50 &	1.17766	&	1.19739	&	1.18706	&	1.19072	&	\bf1.15894	&	-	&	1.20921	\\
VOC2012-ResNet101 &	\bf0.99979	&	1.01592	&	1.04046	&	1.0095	&	1.01902	& -	&	1.18937	\\
\end{tabular}
}
\vspace{3pt}
\vspace{-5pt}
\label{table:feature-correct}
\end{center}
\end{table*}

\section{Experiment}
\label{sec-ex}
To evaluate explanation methods,
we conducted experiments on the CIFAR-10~\citep{krizhevsky2009learning} dataset
and the Pascal VOC 2012~\citep{Everingham10} dataset.
The Pascal VOC 2012 dataset is mainly used for object detection.
Just like in~\citep{interpretableCNN}, we cropped objects using their bounding boxes.
We used the cropped objects as inputs to train DNNs for multi-category classification.
We trained and explained LeNet~\citep{lecun-98}, ResNet-20/32/44/56~\citep{ResNet} using the CIFAR-10 dataset.
AlexNet~\citep{Krizhevsky2012ImageNet},
VGG-16/19~\citep{Simonyan2014Very}, ResNet-50/101~\citep{ResNet} were trained using the
Pascal VOC 2012 dataset.

\subsection{Baseline}
In our experiments, we mainly evaluated the following explanation methods.
Figure~\ref{fig:attribution} shows attribution maps yielded by these explanation methods.\\
\textbf{Grad:} Given an input,~\citep{simonyan2013deep} quantified the attribution value with
the gradient of the input.
We termed this algorithm as Grad.
For RGB images with multiple channels, Grad selected the maximum magnitude across all channels
for each pixel.\\
\textbf{GI:} ~\citep{shrikumar2016not} proposed a method, namely GI,
which used the pixel-wise product of the input and its gradient
as attribution value.
Attribution values for RGB channels were summed to get the final attribution value.\\
\textbf{GB:} Guided Back-propagation, namely GB, corresponded to Grad
where the back-propagation rule at ReLU units was redefined~\citep{springenberg2014striving}.\\
\textbf{LRP:} Layer-wise relevance propagation (LRP)~\citep{binder2016layer}
redefined back-propagation rules for each layer to decompose the output of a DNN over the input.
We used LRP-$\epsilon$ and set the parameter $\epsilon = 1$ in experiments.\\
\textbf{DeepSHAP:} DeepSHAP adapted DeepLIFT~\citep{shrikumar2016not} to approximate pixel-wise Shapley values
for the input image~\citep{lundberg2017unified}.
We applied the code released by~\citep{lundberg2017unified}.\\
\textbf{LIME:} LIME~\citep{ribeiro2016should} trained an interpretable model to compute the attribute value for each super-pixel.
We used the code released by~\citep{ribeiro2016should}.\\
\textbf{Pert:}~\citep{fong2017interpretable} explained a prediction by
training a mask to perturb the input image.
Mask values ranging between 0 and 1 indicated the saliency of each pixel.
We termed this method Pert.\\
\textbf{CAM:} CAM computed attribution map over the feature from the last convolutional layer~\citep{zhou2016learning}.
It required the special structure with a global average pooling layer and a fully connected layer at the end of the DNN.\\
\textbf{Grad\_CAM:} Grad\_CAM was similar to CAM~\citep{selvaraju2017grad}.
Grad\_CAM used gradients over the feature map, instead of the parameters of the fully connected layer.

\subsection{Implementation Details}
\textbf{Bias of the attribution map at the pixel level:}
To approximate the Shapley value for each image,
we sampled 1000 times for each image in the CIFAR-10 dataset
and sampled 100 times for each image in the Pascal VOC 2012 dataset.
We sampled the top-10\%, 30\%, 50\%, 70\%, 90\% pixels with the highest/lowest values.

\textbf{Quantification of unexplainable feature components:}
Given an image, we masked the pixels with the lowest absolute attribution value.
The number of the masked pixels was determined to ensure that
the sum of masked absolute attribution value
took 5\% of the total absolute attribution value.
On average, around 30\% pixels were masked.
The masked pixels were assigned with the average pixel value over images.
We used features of the last convolutional layer to compute $M_{\textrm{feature}}$.

\begin{table*}[!t]
\caption{Non-robustness of explanation methods with different datasets/models.
GB and Grad\_CAM had the lowest non-robustness.
For more results, the supplementary material shows examples of attribution maps of masked images.}
\begin{center}
\resizebox{0.95\linewidth}{!}{\
\begin{small}
\begin{tabular}{l|cccccccc}
Method & Grad & GI & GB & LIME & LRP & Pert & CAM & Grad\_CAM\\
\hline
CIFAR10-LeNet     & 0.503 & 0.608 & 0.473 & 6.799 & 1.139 & \bf 0.472 & - & 1.302\\
CIFAR10-ResNet20  & 0.466 & 0.545 & \bf 0.299 & 1.199 & - & 0.455 & 0.422 & 0.394\\
CIFAR10-ResNet32  & 0.661 & 0.733 & \bf 0.301 & 1.626 & - & 0.458 & 0.399 & 0.400\\
CIFAR10-ResNet44  & 0.809 & 0.866 & \bf 0.269 & 1.239 & - & 0.466 & 0.358 & 0.349\\
CIFAR10-ResNet56  & 0.752 & 0.794 & \bf 0.289 & 3.023 & - & 0.447 & 0.376 & 0.372\\
VOC2012-AlexNet   & 0.439 & 0.519 & \bf 0.359 & 1.698 & 0.654 & 0.678 & - & 0.442\\
VOC2012-VGG16     & 0.362 & 0.382 & \bf 0.239 & 3.597 & 0.378 & 0.607 & - & 0.308\\
VOC2012-VGG19     & 0.384 & 0.398 & \bf 0.256 & 2.074 & 0.393 & 0.653 & - & 0.303\\
VOC2012-ResNet50  & 0.548 & 0.605 & \bf 0.258 & 1.538 & - & 0.518 & 0.290 & 0.308\\
VOC2012-ResNet101 & 0.460 & 0.493 & \bf 0.225 & 1.408 & - & 0.525 & 0.248 & 0.296\\
\end{tabular}
\end{small}}
\end{center}
\label{table:robustness}
\vspace{-10pt}
\end{table*}

\begin{figure*}[!t]
    \centering
    \includegraphics[width=\linewidth]{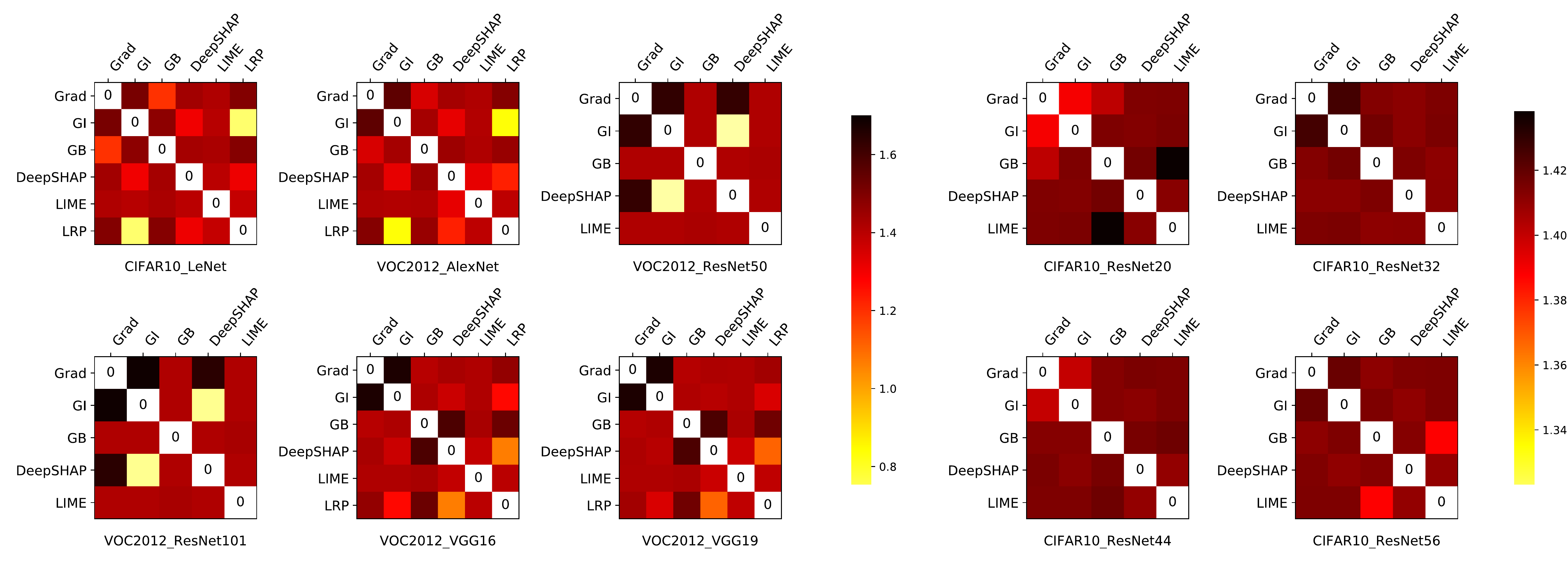}
        \vspace{-9pt}
    \caption{Heat maps of mutual verification.
    A low value of $M_{mutual}$ between two methods indicates a more convincing mutual verification between them,
    \emph{e.g.} LRP and GI had convincing results of mutual verification.
    The supplementary material provides more numerical results.
    }
    \label{fig:mutual-veri}
\end{figure*}

\subsection{Experiment Result and Analysis}
\vspace{-3pt}
\textbf{Bias of the attribution map at the pixel level:}
Figure~\ref{fig:pixel-level} shows curves of evaluation results on different models learned
using different datasets.
According to these curves,
GI and GB provided the least biased attribution maps for ResNet at the pixel level.
For AlexNet, VGG-16/19 and LeNet, LRP outperformed other methods.
Besides, we found that the performance of LIME was volatile, \emph{i.e.} in some cases, LIME
performed quite well, but in some other cases, LIME performed worst.
This situation was obvious on the CIFAR-10 dataset.
We believed that it was because LIME calculated attribution maps for super-pixels.
The number of super-pixels in an image from CIFAR-10 dataset was limited.
In this way, many pixels within a single super-pixel shared the same attribution value in results of LIME,
which made it hard to sample these pixels with significantly biased attribution values.

Some methods could not be evaluated using the bias at the pixel level.
For example, Pert computed an importance map without negative values
instead of an attribution map for each image.
The code of CAM~\citep{zhou2016learning} projected attribution
values to the range between 0 and 1.
Grad\_CAM and LRP were not used on residual networks.
Because there was only one fully connected layer behind the last convolutional layer in residual networks,
in this case, Grad\_CAM could not reflect the information processing contained in the cascaded non-liner layers
of the DNN.
For LRP, the relevance propagation rules of some structures in ResNet were not defined
to the best of our knowledge.

\textbf{Quantification of unexplainable feature components:}
Table~\ref{table:feature-correct} compares the amount of unexplainable feature components
between explanation methods.
We found that LIME, GB and Pert explained more feature components than other methods.
We noticed that the quantification of unexplainable feature components of most explanation methods
were considerable larger than expected.
It was because the attribution maps from some methods contained relatively larger noise.
Thus, the masked pixels were almost uniformly distributed over images,
which destroyed the context information and led to worse results.

We did not evaluate CAM and Grad\_CAM, because they calculated attribution maps
at the feature level,
which were not comparable with attribution values at the pixel level.

\textbf{Robustness of explanation:}
Table~\ref{table:robustness} shows the quantitative results of
$M_{\textrm{non-robust}}$ on different models trained using different datasets.
We found that GB and Grad\_CAM exhibited a lower non-robustness to spatial masking.
LIME segmented the input image into super pixels to calculate attributions,
and the spatial masking could influence the segmentation result significantly.
Thus, non-robustness of LIME was much higher than other explanation methods.

\textbf{Mutual verification:}
Figure~\ref{fig:mutual-veri} visualizes the mutual verification $M_{\textrm{mutual}}$
between different explanation methods,
which indicates a high level mutual verification between LRP, GI and DeepSHAP.
Note that we did not compare CAM and Grad\_CAM with other methods.
It was because they computed attribution maps on intermediate-layer features.

\section{Conclusion}
\vspace{-5pt}
\label{sec-conclu}
In this paper, we have proposed four metrics to evaluate
explanation methods from four different perspectives.
The proposed evaluation metrics are computed without requirements for ground-truth explanations.
Our metrics can be applied to widely used explanation methods \emph{w.r.t.}
different DNNs learned using different datasets.
These metrics evaluate the bias of the attribution map at the pixel level,
quantify the unexplainable feature components,
the robustness of the explanation and the mutual verification.
In experiments,
we used our metrics to evaluate nine widely used explanation methods.
Experimental results showed that attribution maps from LRP, GI and GB exhibited lower bias at the pixel level.
LIME and GB explained more feature components than other methods.
Regarding the robustness,
GB, CAM and Grad\_CAM were more robust to spatial masking than other explanation methods.
DeepSHAP, GI and LRP can better verified each other.

\bibliography{paper}
\bibliographystyle{iclr2020_conference}

\newpage
\appendix
\section{Properties of the Shapley value}
\label{sec:shap-prop}
Let $I$ denote the input image; let $\Omega$ denote the set of all pixels in $I$.
We can use $I_\emptyset$ to denote a baseline image,
\emph{i.e.} all pixels in $I_\emptyset$ equal to the average value over all images.
For a subset $S\subset \Omega$,
$I_S$ denotes an image that satisfies
\begin{small}
\vspace{-2pt}
\begin{equation}
    (I_S)_i=\begin{cases}
                (I)_i, & i \in S\\
                (I_\emptyset)_i, & i \notin S
\end{cases}
\vspace{-2pt}
\end{equation}
\end{small}

where $i$ is the index of the pixel in $I$ and $I_\Omega$ is the same image as $I$.
Let $F$ and $G$ denote two models with scalar output.
The Shapley value of the $i$-th pixel is represented by $A^*_i$, and they
have the following properties~\cite{shapley1953value}.

\textbf{Efficiency:} The sum of Shapley values $\sum_{i\in \Omega}A^{*}_i$ is equal to $F(I_\Omega)-F(I_\emptyset)$.

\textbf{Symmetry:} The features that are treated equally by the model are treated equally by the Shapley value.
If $F(I_{S\cup\{i\}})=F(I_{S\cup\{j\}})$ for all subsets S, then $A^{*}_i=A^{*}_j$.

\textbf{Additivity:}  For any two models $F$ and $G$, if they are combined into one model $F+G$,
the Shapley value must be added pixel by pixel: $(A^*)^{F+G}_i=(A^*)^F_i+(A^*)^G_i$.

\textbf{Monotonicity:}  For any two models $F$ and $G$, if for all subsets $S$ we have
$F(I_{S\cup\{i\}})-F(I_S)\ge G(I_{S\cup \{i\}})-G(I_S)$ for all subsets S,
then we have $(A^{*})^F_i\ge (A^{*})^G_i$.

\section{Analysis of the computational cost}
\label{sec:analysis}
In this section, we continue using the notation in Section 3.1 and Section 3.2.
Suppose that we sample $m$ times to approximate the Shapley value.
The variance of $A^{shap}_i$ is $\frac{\sigma^2}{m}$ where $\sigma^2$ satisfies~\cite{castro2009polynomial}
\begin{small}
\vspace{-2pt}
\begin{equation}
    \sigma^2=\sum_{P\subset \Omega\setminus \{i\}}\frac{|P|!(|\Omega|-|P|-1)!}{|P|!}
    \left[F(I_{P\cup \{i\}})- F(I_P)-A^{*}_i\right]^2
\vspace{-2pt}
\end{equation}
\end{small}

So we have $(\sigma^{shap})^2=\sigma^2/m$.
For the set of sampled pixels $S$,
the variance of their average Shapley value is
{\small $\frac{|S|(\sigma^{shap})^2}{|S|^2} = \frac{(\sigma^{shap)^2}}{|S|} = \frac{\sigma^2}{m|S|}$}.
Apparently, if we want to get the same accuracy for a single pixel as the set of pixels,
we need to sample $m|S|$ times,
which needs much more computational cost than our metric, especially when the number of sampled pixels is large.

\newpage
\section{More examples of attribution maps}
\label{sec:more-attribution-maps}

\begin{figure*}[h]
    \centering
    \includegraphics[width=0.9\linewidth]{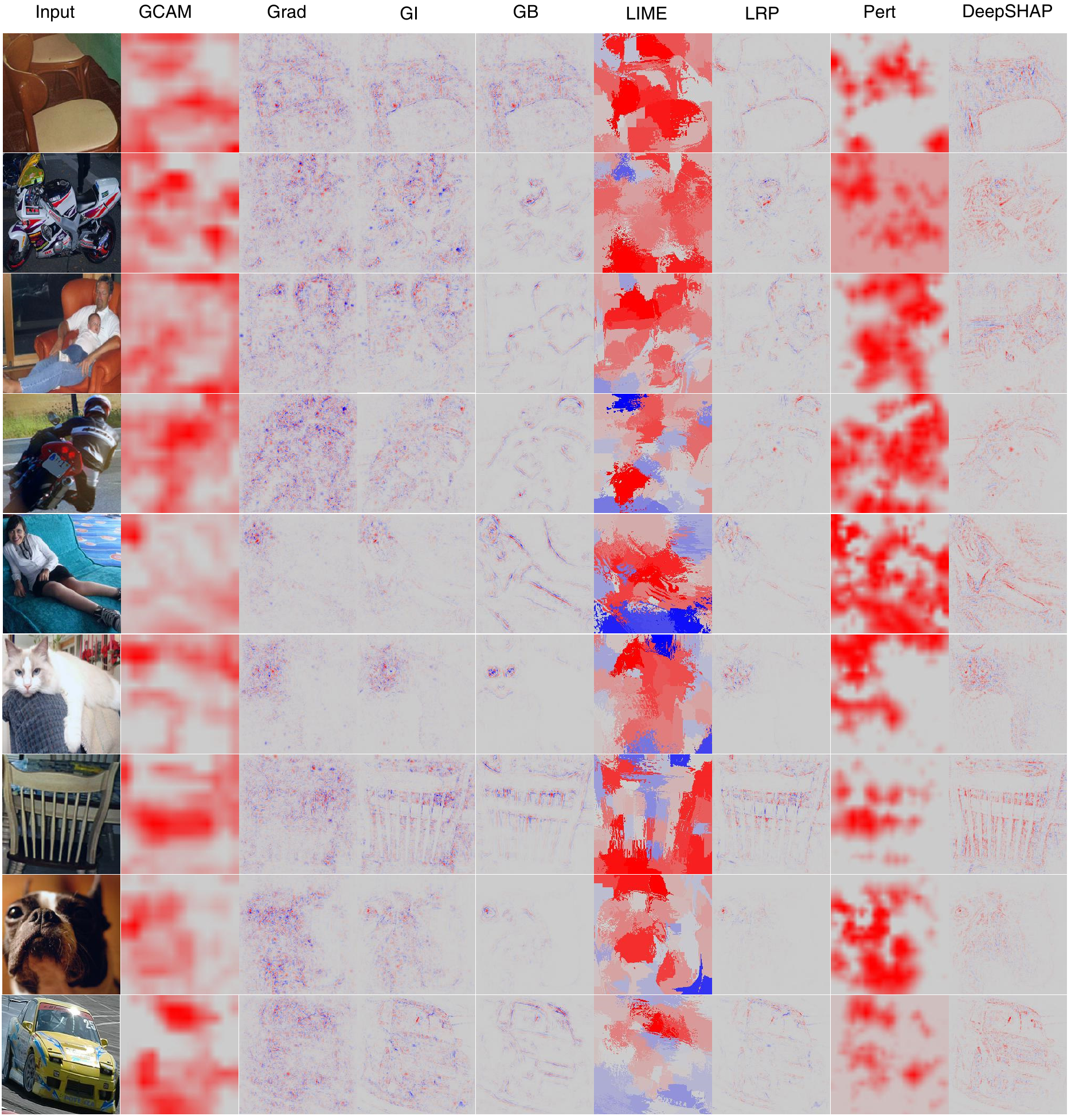}
    \caption{Examples of attribution maps.}
    \label{fig:more-attribution}
\end{figure*}

\newpage
\section{Detailed results of the bias of the attribution map at the pixel level}

\label{sec:more-result-1}
\begin{table*}[h]
\caption{Bias of the attribution map at the pixel level on CIFAR-10-LeNet}
\label{table:pixel-1}
\begin{center}
\begin{small}
\begin{tabular}{l|ccccccc}
Method & Grad\_CAM & \big|Grad\big| & GI & GB & DeepSHAP & LIME & LRP\\
\hline
top-10\% & 0.17256& 0.05302& 0.04060& 0.04701& 0.04924& \bf 0.03138& 0.04164\\
top-30\% & 0.04921& 0.03090& 0.01892& 0.02109& 0.02408& 0.03051&\bf 0.01798\\
top-50\% & 0.01961& 0.02057& 0.01126& 0.01225& 0.01443& 0.02919&\bf 0.01041\\
top-70\% & 0.01531& 0.01137& 0.00671& 0.00746& 0.00902& 0.02727&\bf 0.00634\\
top-90\% & 0.01785& 0.00314& \bf 0.00240& 0.00300& 0.00334& 0.02537& 0.00277\\
bottom-10\% & 0.05317& 0.06326& 0.04907& 0.06730& 0.05755& \bf 0.02160& 0.04271\\
bottom-30\% & 0.04510& 0.03948& 0.02626& 0.03224& 0.03100& \bf 0.02150& 0.02198\\
bottom-50\% & 0.04732& 0.02840& 0.01767& 0.02126& 0.02047& 0.02222& \bf 0.01481\\
bottom-70\% & 0.04693& 0.01886& 0.01270& 0.01551& 0.01462& 0.02286& \bf 0.01086\\
bottom-90\% & 0.04022& 0.01026& 0.00808& 0.01025& 0.00882& 0.02385& \bf 0.00708\\
\end{tabular}
\end{small}
\end{center}
\end{table*}

\begin{table*}[h]
\caption{Bias of the attribution map at the pixel level on CIFAR-10-ResNet20}
\label{table:pixel-2}
\begin{center}
\begin{small}
\begin{tabular}{l|ccccc}
Method   & \big|Grad\big| & GI & GB & DeepSHAP & LIME \\
\hline
top-10\% & 0.04692& 0.03962& 0.04801& 0.04702& \bf 0.02716\\
top-30\% & 0.02552& \bf 0.01522& 0.01848& 0.02175& 0.02706\\
top-50\% & 0.01616& \bf 0.00795& 0.00943& 0.01241& 0.02666\\
top-70\% & 0.00762& \bf 0.00416& 0.00425& 0.00643& 0.02456\\
top-90\% & 0.00288& 0.00283& 0.00336& \bf 0.00270& 0.02185\\
bottom-10\% & 0.06610& 0.06449& 0.06726& 0.06866& \bf 0.01790\\
bottom-30\% & 0.04272& 0.03437& 0.03833& 0.03962& \bf 0.01765\\
bottom-50\% & 0.03117& 0.02296& 0.02704& 0.02722& \bf 0.01836\\
bottom-70\% & 0.02168& 0.01726& 0.02053& 0.01992& \bf 0.01897\\
bottom-90\% & 0.01357& \bf 0.01275& 0.01513& 0.01346& 0.02005\\
\end{tabular}
\end{small}
\end{center}
\end{table*}

\begin{table*}[h]
\caption{Bias of the attribution map at the pixel level on CIFAR-10-ResNet32}
\label{table:pixel-3}
\begin{center}
\begin{small}
\begin{tabular}{l|ccccc}
Method   & \big|Grad\big| & GI & GB & DeepSHAP & LIME \\
\hline
top-10\% & 0.04836& 0.04126& 0.03984& 0.04773& \bf 0.02699\\
top-30\% & 0.02600& \bf 0.01641& 0.01728& 0.02247& 0.02702\\
top-50\% & 0.01639& \bf 0.00874& 0.00972& 0.01289& 0.02630\\
top-70\% & 0.00793& \bf 0.00459& 0.00565& 0.00683& 0.02465\\
top-90\% & 0.00268& 0.00264& \bf 0.00242& 0.00260& 0.02267\\
bottom-10\% & 0.06535& 0.06389& 0.07271& 0.06787& \bf 0.02059\\
bottom-30\% & 0.04217& 0.03433& 0.03612& 0.03917& \bf 0.02053\\
bottom-50\% & 0.03064& 0.02304& 0.02359& 0.02686& \bf 0.02065\\
bottom-70\% & 0.02135& \bf 0.01726& 0.01733& 0.01963& 0.02103\\
bottom-90\% & 0.01331& 0.01254& \bf 0.01214& 0.01307& 0.02145\\
\end{tabular}
\end{small}
\end{center}
\end{table*}

\begin{table*}[h]
\caption{Bias of the attribution map at the pixel level on CIFAR-10-ResNet44}
\label{table:pixel-4}
\begin{center}
\begin{small}
\begin{tabular}{l|ccccc}
Method   & \big|Grad\big| & GI & GB & DeepSHAP & LIME \\
\hline
top-10\% & 0.04795& 0.04155& 0.04055& 0.04751& \bf 0.02712\\
top-30\% & 0.02576& \bf 0.01626& 0.01698& 0.02254& 0.02709\\
top-50\% & 0.01609& \bf 0.00848& 0.00935& 0.01307& 0.02637\\
top-70\% & 0.00748& \bf 0.00433& 0.00502& 0.00686& 0.02485\\
top-90\% & 0.00248& 0.00228& 0.00230& \bf 0.00226& 0.02271\\
bottom-10\% & 0.06510& 0.06452& 0.07143& 0.06802& \bf 0.01924\\
bottom-30\% & 0.04196& 0.03449& 0.03663& 0.03977& \bf 0.01879\\
bottom-50\% & 0.03082& 0.02323& 0.02448& 0.02736& \bf 0.01967\\
bottom-70\% & 0.02159& \bf 0.01752& 0.01811& 0.01989& 0.02064\\
bottom-90\% & 0.01353& \bf 0.01282& 0.01293& 0.01323& 0.02137\\

\end{tabular}
\end{small}
\end{center}
\end{table*}

\begin{table*}[h]
    \caption{Bias of the attribution map at the pixel level on CIFAR-10-ResNet56}
    \label{table:pixel-5}
    \begin{center}
    \begin{small}
    \begin{tabular}{l|ccccc}
    Method   & \big|Grad\big| & GI & GB & DeepSHAP & LIME \\
    \hline
    top-10\% & 0.04716& 0.04129& 0.04780& 0.04601& \bf 0.02690\\
    top-30\% & 0.02537& \bf 0.01659& 0.01855& 0.02089& 0.02714\\
    top-50\% & 0.01607& \bf 0.00900& 0.01022& 0.01215& 0.02679\\
    top-70\% & 0.00791& \bf 0.00482& 0.00569& 0.00652& 0.02509\\
    top-90\% & 0.00212& \bf 0.00194& 0.00214& 0.00234& 0.02252\\
    bottom-10\% & 0.06625& 0.06551& 0.06065& 0.06727& \bf 0.01828\\
    bottom-30\% & 0.04164& 0.03449& 0.03158& 0.03779& \bf 0.01812\\
    bottom-50\% & 0.03001& 0.02299& 0.02129& 0.02574& \bf 0.01911\\
    bottom-70\% & 0.02086& \bf 0.01712& 0.01587& 0.01868& 0.02023\\
    bottom-90\% & 0.01300& 0.01237& \bf 0.01146& 0.01267& 0.02088\\

    \end{tabular}
    \end{small}
    \end{center}
    \end{table*}

\begin{table*}[h]
\caption{Bias of the attribution map at the pixel level on VOC2012-ResNet50}
\label{table:pixel-6}
\begin{center}
\begin{small}
\begin{tabular}{l|ccccc}
Method   & \big|Grad\big| & GI & GB & DeepSHAP & LIME \\
\hline
top-10\% & 0.00766& 0.00728& \bf 0.00633& 0.00738& 0.00876\\
top-30\% & 0.00437& 0.00345& \bf 0.00276& 0.00359& 0.00606\\
top-50\% & 0.00297& 0.00213& \bf 0.00170& 0.00223& 0.00451\\
top-70\% & 0.00173& 0.00138& \bf 0.00113& 0.00144& 0.00335\\
top-90\% & 0.00060& 0.00060& \bf 0.00053& 0.00062& 0.00235\\
bottom-10\% & 0.00831& 0.00824& 0.00850& 0.00839& \bf 0.00430\\
bottom-30\% & 0.00500& 0.00417& 0.00386& 0.00432& \bf 0.00240\\
bottom-50\% & 0.00354& 0.00269& 0.00244& 0.00280& \bf 0.00151\\
bottom-70\% & 0.00228& 0.00189& 0.00171& 0.00194& \bf 0.00113\\
bottom-90\% & 0.00117& 0.00112& \bf 0.00111& 0.00114& 0.00128\\

\end{tabular}
\end{small}
\end{center}
\end{table*}

\begin{table*}[h]
\caption{Bias of the attribution map at the pixel level on VOC2012-ResNet101}
\label{table:pixel-7}
\begin{center}
\begin{small}
\begin{tabular}{l|ccccc}
Method   & \big|Grad\big| & GI & GB & DeepSHAP & LIME \\
\hline
top-10\% & 0.00790& 0.00678& \bf 0.00610& 0.00685& 0.00827\\
top-30\% & 0.00438& 0.00307& \bf 0.00260& 0.00318& 0.00608\\
top-50\% & 0.00291& 0.00185& \bf 0.00158& 0.00194& 0.00473\\
top-70\% & 0.00175& 0.00121& \bf 0.00104& 0.00127& 0.00371\\
top-90\% & 0.00067& 0.00055& \bf 0.00049& 0.00056& 0.00278\\
bottom-10\% & 0.00861& 0.00754& 0.00839& 0.00767& \bf 0.00387\\
bottom-30\% & 0.00496& 0.00369& 0.00377& 0.00381& \bf 0.00222\\
bottom-50\% & 0.00343& 0.00237& 0.00238& 0.00245& \bf 0.00142\\
bottom-70\% & 0.00225& 0.00169& 0.00169& 0.00173& \bf 0.00124\\
bottom-90\% & 0.00117& \bf 0.00104& 0.00112& 0.00105& 0.00174\\

\end{tabular}
\end{small}
\end{center}
\end{table*}

\begin{table*}[h]
\caption{Bias of the attribution map at the pixel level on VOC2012-AlexNet}
\label{table:pixel-8}
\begin{center}
\begin{small}
\begin{tabular}{l|ccccccc}
Method & Grad\_CAM & \big|Grad\big| & GI & GB & DeepSHAP & LIME & LRP\\
\hline
top-10\% & 0.05403& 0.00744& 0.00505& 0.00646& 0.00845& 0.00854& \bf 0.00453\\
top-30\% & 0.03467& 0.00434& 0.00243& 0.00296& 0.00447& 0.00598& \bf 0.00197\\
top-50\% & 0.02459& 0.00298& 0.00149& 0.00182& 0.00295& 0.00443& \bf 0.00119\\
top-70\% & 0.01711& 0.00176& 0.00094& 0.00120& 0.00209& 0.00329& \bf 0.00080\\
top-90\% & 0.01115& 0.00060& \bf 0.00035& 0.00058& 0.00140& 0.00222& 0.00043\\
bottom-10\% & 0.01026& 0.00906& 0.00638& 0.00876& 0.00438& 0.00466& \bf 0.00430\\
bottom-30\% & 0.00978& 0.00534& 0.00327& 0.00400& 0.00215& 0.00267& \bf 0.00202\\
bottom-50\% & 0.00803& 0.00372& 0.00215& 0.00253& 0.00131& 0.00165& \bf 0.00129\\
bottom-70\% & 0.00610& 0.00239& 0.00151& 0.00178& \bf 0.00075& 0.00103& 0.00091\\
bottom-90\% & 0.00690& 0.00124& 0.00093& 0.00111& \bf 0.00030& 0.00104& 0.00056\\

\end{tabular}
\end{small}
\end{center}
\end{table*}

\begin{table*}[h]
\caption{Bias of the attribution map at the pixel level on VOC2012-VGG16}
\label{table:pixel-9}
\begin{center}
\begin{small}
\begin{tabular}{l|ccccccc}
Method & Grad\_CAM & \big|Grad\big| & GI & GB & DeepSHAP & LIME & LRP\\
\hline
top-10\% & 0.02984& 0.00764& 0.00656& 0.00563& 0.00768& 0.00864& \bf 0.00461\\
top-30\% & 0.01876& 0.00405& 0.00292& 0.00253& 0.00310& 0.00554& \bf 0.00159\\
top-50\% & 0.01365& 0.00267& 0.00176& 0.00154& 0.00187& 0.00387& \bf 0.00092\\
top-70\% & 0.00947& 0.00162& 0.00115& 0.00098& 0.00132& 0.00272& \bf 0.00065\\
top-90\% & 0.00650& 0.00061& 0.00051& \bf 0.00038& 0.00089& 0.00172& 0.00044\\
bottom-10\% & 0.00767& 0.00869& 0.00768& 0.00865& 0.00396& 0.00580& \bf 0.00288\\
bottom-30\% & 0.00708& 0.00483& 0.00372& 0.00406& 0.00173& 0.00344& \bf 0.00117\\
bottom-50\% & 0.00654& 0.00330& 0.00238& 0.00261& 0.00106& 0.00223& \bf 0.00072\\
bottom-70\% & 0.00540& 0.00219& 0.00170& 0.00185& 0.00075& 0.00153& \bf 0.00054\\
bottom-90\% & 0.00509& 0.00120& 0.00107& 0.00122& 0.00041& 0.00114& \bf 0.00040\\

\end{tabular}
\end{small}
\end{center}
\end{table*}

\begin{table*}[h]
\caption{Bias of the attribution map at the pixel level on VOC2012-VGG19}
\label{table:pixel-10}
\begin{center}
\begin{small}
\begin{tabular}{l|ccccccc}
Method & Grad\_CAM & \big|Grad\big| & GI & GB & DeepSHAP & LIME & LRP\\
\hline
top-10\% & 0.02564& 0.00761& 0.00672& 0.00580& 0.00802& 0.00851& \bf 0.00468\\
top-30\% & 0.01991& 0.00403& 0.00301& 0.00261& 0.00337& 0.00565& \bf 0.00157\\
top-50\% & 0.01517& 0.00263& 0.00181& 0.00159& 0.00205& 0.00402& \bf 0.00091\\
top-70\% & 0.01139& 0.00161& 0.00118& 0.00102& 0.00143& 0.00282& \bf 0.00064\\
top-90\% & 0.00799& 0.00062& 0.00053& \bf 0.00043& 0.00096& 0.00183& 0.00045\\
bottom-10\% & 0.00932& 0.00867& 0.00782& 0.00871& 0.00442& 0.00587& \bf 0.00236\\
bottom-30\% & 0.00764& 0.00480& 0.00378& 0.00406& 0.00193& 0.00352& \bf 0.00095\\
bottom-50\% & 0.00683& 0.00326& 0.00243& 0.00260& 0.00120& 0.00231& \bf 0.00059\\
bottom-70\% & 0.00630& 0.00217& 0.00173& 0.00184& 0.00084& 0.00140& \bf 0.00046\\
bottom-90\% & 0.00649& 0.00119& 0.00109& 0.00120& 0.00043& 0.00107& \bf 0.00036\\
\end{tabular}
\end{small}
\end{center}
\end{table*}

\clearpage
\section{More results of the quantification of unexplainable feature components}
The following table provides the result of the quantification of unexplainable feature components with $\tau=0.1$.

\begin{table*}[h]
\caption{Quantification of unexplainable feature components with $\tau=0.1$.}
\begin{center}
\resizebox{0.9\linewidth}{!}{\
\begin{small}
\centering
\begin{tabular}{l|ccccccc}
Method & Grad& GI & GB & DeepSHAP & LIME & LRP & Pert \\
\hline
CIFAR10-LeNet &	0.96821	&	1.10399	&	0.91546	&	1.10371	& \bf0.67032	&	1.08469	&   0.70177	\\
CIFAR10-ResNet20 &	1.22212	&	1.28065	&	1.14596	&	1.26002	&	\bf1.05286	&	-	&	1.13032\\
CIFAR10-ResNet32 &	1.22803	&	1.29324	&	1.16028	&	1.27681	&	\bf1.04412	&	-	&	1.14261\\
CIFAR10-ResNet44&	1.22382	&	1.26434	&	1.10411	&	1.24961	&	\bf1.02039	&	-	&	1.10772\\
CIFAR10-ResNet56 &	1.22487	&	1.24123	&	1.11306	&	1.24146	&	\bf1.00916	&	-	&	1.10122\\
VOC2012-AlexNet &	1.02425	&	1.09375	&	1.04981	&	1.0809	&	\bf1.01942	&	1.11513	&	1.04048	\\
VOC2012-VGG16 & 1.18959	&	1.22715	&	1.22724	&	1.32364	&	\bf1.12339	&	1.32674	&	1.12878 \\
VOC2012-VGG19 &	\bf1.08084	&	1.1226	&	1.10411	&	1.17121	&	1.08387	&	1.23784	&	1.13043	\\
VOC2012-ResNet50 &	1.23345	&	1.24123	&	\bf1.2156	&	1.23441	&	1.22411	&	-	&	1.25732	\\
VOC2012-ResNet101 &	1.08592	&	1.09363	&	\bf1.07853	&	1.07856	&	1.11289	& -	&	1.23279	\\
\end{tabular}
\end{small}}
\end{center}
\end{table*}

\section{Attribution maps of masked images}
\label{sec:masked-attribution}
\begin{figure*}[h]
    \centering
    \includegraphics[width=0.9\linewidth]{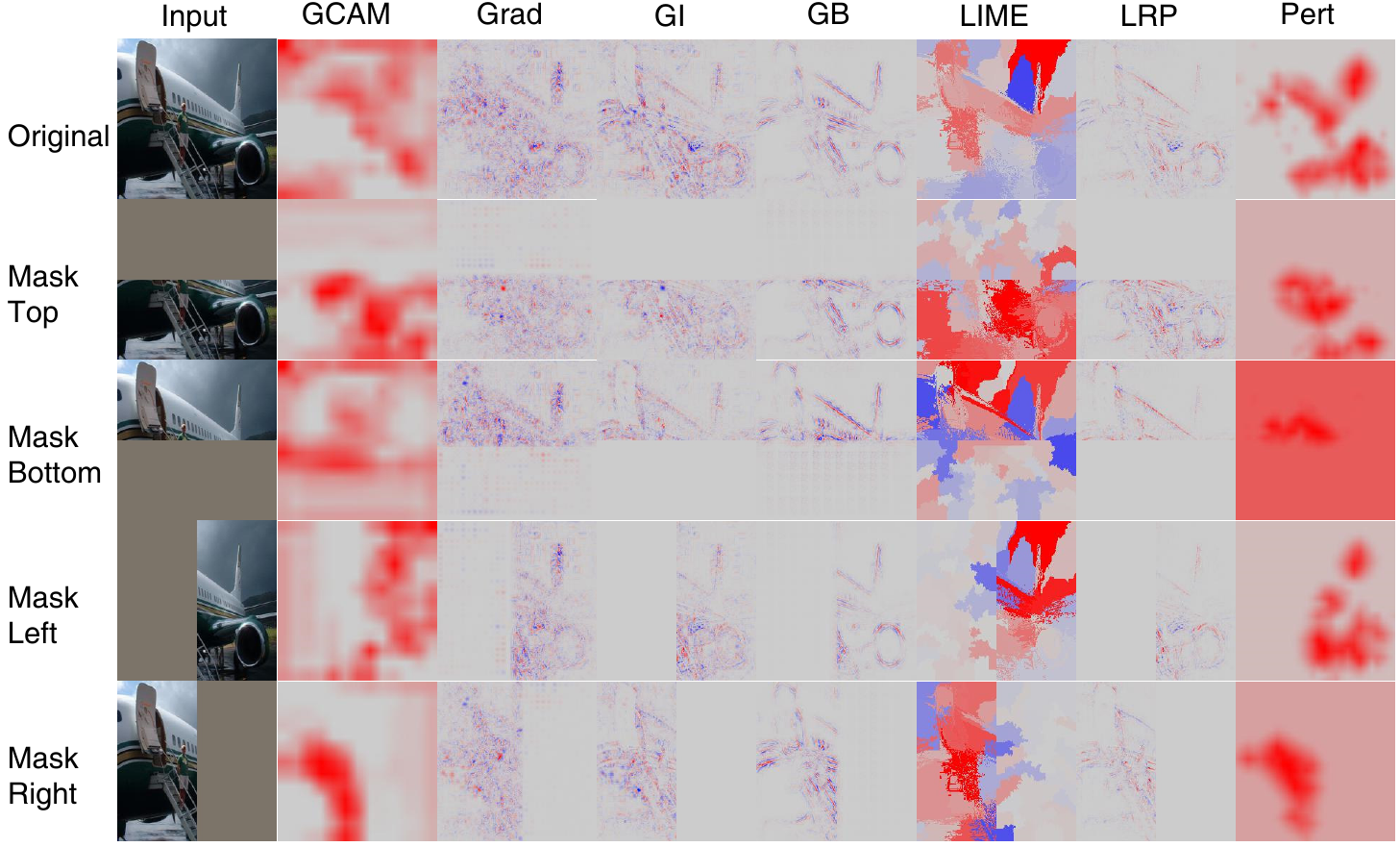}
    \caption{Examples of attribution maps after spatial masking.}
    \label{fig:more-robust}
\end{figure*}

\section{Detailed results of the mutual verification}
\label{sec:more-result-2}

\begin{table*}[!h]
\caption{Mutual verification on VOC2012-VGG19}
\begin{center}
\begin{small}
\begin{tabular}{l|cccccc}
Method & Grad & GI & GB & DeepSHAP & LIME & LRP \\
\hline
Grad & 0.0000& 1.6661& 1.4054& 1.4177& 1.4142& 1.4375\\
GI & 1.6661& 0.0000& 1.4159& 1.4008& 1.4140&  1.3437\\
GB & 1.4054& 1.4159& 0.0000& 1.5834& 1.4244& 1.5242\\
DeepSHAP & 1.4177& 1.4009& 1.5835& 0.0000& 1.3691& \bf 1.1106\\
LIME &1.4142& 1.4141& 1.4245& 1.3691& 0.0000& 1.3874\\
LRP & 1.4375& 1.3437& 1.5243& \bf 1.1106& 1.3874& 0.0000\\
\end{tabular}
\end{small}
\end{center}
\end{table*}

\begin{table*}[!h]
\caption{Mutual verification on VOC2012-VGG16}
\begin{center}
\begin{small}
\begin{tabular}{l|cccccc}
Method & Grad & GI & GB & DeepSHAP & LIME & LRP \\
\hline
Grad & 0.0000& 1.6647& 1.3992& 1.4302& 1.4142& 1.4621\\
GI & 1.6647& 0.0000& 1.4191& 1.3711& 1.4139& 1.2649\\
GB & 1.3992& 1.4191& 0.0000& 1.5839& 1.4260& 1.5320\\
DeepSHAP & 1.4302& 1.3712& 1.5840& 0.0000& 1.3824& \bf 1.0636\\
LIME & 1.4142& 1.4140& 1.4261& 1.3825& 0.0000& 1.3960\\
LRP & 1.4621& 1.2649& 1.5321& \bf 1.0636& 1.3960& 0.0000\\
\end{tabular}
\end{small}
\end{center}
\end{table*}

\begin{table*}[!h]
\caption{Mutual verification on VOC2012-AlexNet}
\begin{center}
\begin{small}
\begin{tabular}{l|cccccc}
Method & Grad & GI & GB & DeepSHAP & LIME & LRP \\
\hline
Grad & 0.0000& 1.5540& 1.3477& 1.4348& 1.4148& 1.4851\\
GI & 1.5541& 0.0000& 1.4344& 1.3224& 1.4113& \bf 0.8341\\
GB & 1.3477& 1.4344& 0.0000& 1.4477& 1.4146& 1.4573\\
DeepSHAP & 1.4349& 1.3225& 1.4478& 0.0000& 1.3221& 1.2251\\
LIME & 1.4149& 1.4115& 1.4147& 1.3222& 0.0000& 1.3916\\
LRP & 1.4851& \bf 0.8341& 1.4573& 1.2250& 1.3915& 0.0000\\
\end{tabular}
\end{small}
\end{center}
\end{table*}

\begin{table*}[!h]
\caption{Mutual verification on VOC2012-ResNet50}
\begin{center}
\begin{small}
\begin{tabular}{l|ccccc}
Method & Grad & GI & GB & DeepSHAP & LIME \\
\hline
Grad & 0.0000& 1.6285& 1.4143& 1.6242& 1.4143\\
GI & 1.6285& 0.0000& 1.4137& \bf 0.6537& 1.4139\\
GB & 1.4143& 1.4136& 0.0000& 1.4137& 1.4238\\
DeepSHAP & 1.6242& \bf 0.6537& 1.4137& 0.0000& 1.4139\\
LIME & 1.4144& 1.4141& 1.4240& 1.4140& 0.0000\\
\end{tabular}
\end{small}
\end{center}
\end{table*}

\begin{table*}[!h]
\caption{Mutual verification on VOC2012-ResNet101}
\begin{center}
\begin{small}
\begin{tabular}{l|ccccc}
Method & Grad & GI & GB & DeepSHAP & LIME \\
\hline
Grad & 0.0000& 1.6877& 1.4136& 1.6428& 1.4145\\
GI & 1.6877& 0.0000& 1.4140& \bf 0.6778& 1.4139\\
GB & 1.4136& 1.4140& 0.0000& 1.4140& 1.4275\\
DeepSHAP & 1.6428& \bf 0.6778& 1.4140& 0.0000& 1.4138\\
LIME & 1.4146& 1.4140& 1.4277& 1.4139& 0.0000\\
\end{tabular}
\end{small}
\end{center}
\end{table*}

\begin{table*}[!h]
\caption{Mutual verification on CIFAR10-ResNet56}
\begin{center}
\begin{small}
\begin{tabular}{l|ccccc}
Method & Grad & GI & GB & DeepSHAP & LIME \\
\hline
Grad & 0.0000& 1.4183& 1.4107& 1.4135& 1.4142\\
GI & 1.4183& 0.0000& 1.4140& 1.4104& 1.4140\\
GB & 1.4107& 1.4140& 0.0000& 1.4128& 1.3879\\
DeepSHAP & 1.4135& 1.4104& 1.4128& 0.0000& 1.4097\\
LIME & 1.4142& 1.4140& 1.3879& 1.4098& 0.0000\\
\end{tabular}
\end{small}
\end{center}
\end{table*}

\clearpage

\begin{table*}[!h]
\caption{Mutual verification on CIFAR10-ResNet44}
\begin{center}
\begin{small}
\begin{tabular}{l|ccccc}
Method & Grad & GI & GB & DeepSHAP & LIME \\
\hline
Grad & 0.0000& 1.3999& 1.4128& 1.4147& 1.4141\\
GI & 1.3999& 0.0000& 1.4125& 1.4114& 1.4143\\
GB & 1.4128& 1.4125& 0.0000& 1.4152& 1.4175\\
DeepSHAP & 1.4147& 1.4114& 1.4152& 0.0000& 1.4097\\
LIME & 1.4142& 1.4144& 1.4152& 1.4175& 0.0000\\
\end{tabular}
\end{small}
\end{center}
\end{table*}

\begin{table*}[!h]
\caption{Mutual verification on CIFAR10-ResNet32}
\begin{center}
\begin{small}
\begin{tabular}{l|ccccc}
Method & Grad & GI & GB & DeepSHAP & LIME \\
\hline
Grad & 0.0000& 1.4259& 1.4129& 1.4117& 1.4143\\
GI & 1.4259& 0.0000& 1.4164& 1.4113& 1.4145\\
GB & 1.4129& 1.4164& 0.0000& 1.4144& 1.4111\\
DeepSHAP & 1.4117& 1.4113& 1.4144& 0.0000& 1.4117\\
LIME & 1.4144& 1.4145& 1.4112& 1.4117& 0.0000\\
\end{tabular}
\end{small}
\end{center}
\end{table*}

\begin{table*}[!h]
\caption{Mutual verification on CIFAR10-ResNet20}
\begin{center}
\begin{small}
\begin{tabular}{l|ccccc}
Method & Grad & GI & GB & DeepSHAP & LIME \\
\hline
Grad & 0.0000& 1.3895& 1.4010& 1.4134& 1.4143\\
GI & 1.3895& 0.0000& 1.4140& 1.4133& 1.4145\\
GB & 1.4010& 1.4140& 0.0000& 1.4164& 1.4382\\
DeepSHAP & 1.4134& 1.4133& 1.4164& 0.0000& 1.4118\\
LIME & 1.4144& 1.4145& 1.4382& 1.4118& 0.0000\\
\end{tabular}
\end{small}
\end{center}
\end{table*}

\begin{table*}[!h]
\caption{Mutual verification on CIFAR10-LeNet}
\begin{center}

\begin{small}
\begin{tabular}{l|cccccc}
Method & Grad & GI & GB & DeepSHAP & LIME & LRP \\
\hline
Grad & 0.0000& 1.5079& 1.1955& 1.4352& 1.4130& 1.4904\\
GI & 1.5079& 0.0000& 1.4714& 1.3025& 1.4003& \bf 0.7179\\
GB & 1.1955& 1.4714& 0.0000& 1.4329& 1.4249& 1.4881\\
DeepSHAP & 1.4352& 1.3025& 1.4329& 0.0000& 1.3958& 1.3063\\
LIME & 1.4131& 1.4004& 1.4249& 1.3959& 0.0000& 1.3805\\
LRP & 1.4904& \bf 0.7179& 1.4881& 1.3063& 1.3804& 0.0000\\
\end{tabular}
\end{small}
\end{center}
\end{table*}

\end{document}